\documentclass{article}
\usepackage{spconf,amsmath,graphicx}
\usepackage[french]{babel}
\usepackage[T1]{fontenc}
\usepackage[utf8]{inputenc}
\usepackage{multirow}
\usepackage{float}
\usepackage{adjustbox}
\usepackage{multirow}

\usepackage{color,array,dcolumn}
\usepackage{booktabs,amsmath}

\title{Bilinear residual Neural Network for the identification and forecasting of dynamical systems}
\name{Ronan Fablet $^1$, Said Ouala $^1$, Cédric Herzet $^{1,2}$ \thanks{This work was supported by Labex Cominlabs (grant SEACS) and CNES (grant OSTST-MANATEE).}}
\address{$(1)$ IMT Atlantique; Lab-STICC, Brest, France\\
$(2)$ INRIA Bretagne-Atlantique, Fluminance, Rennes, France}
\begin{document}
%
\maketitle
%
%
\begin{abstract} 
Due to the increasing availability of large-scale observation and simulation datasets, data-driven representations arise as efficient and relevant computation representations of dynamical systems for a wide range of applications, where model-driven models based on ordinary differential equation remain the state-of-the-art approaches. In this work, we investigate neural networks (NN) as physically-sound data-driven representations of such systems. Reinterpreting Runge-Kutta methods as graphical models, we consider a residual NN architecture and introduce bilinear layers to embed non-linearities which are intrinsic features of dynamical systems. From numerical experiments for classic dynamical systems, we demonstrate the relevance of the proposed NN-based architecture both in terms of forecasting performance and model identification.    
\end{abstract}
\section{Problem statement and related work}
\label{sec:intro}

Model-driven strategies have long been the classic framework to address forecasting and reconstruction of physical systems. Typical examples may be taken from geosciences \cite{evensen_data_2009}. The ever increasing availability of large-scale observation and simulation datasets make more and more appealing the development of data-driven strategies especially when dealing with computationally-demanding models or high modeling uncertainties \cite{evensen_data_2009}.

In this context, data-driven schemes typically aim to identify computational representations of the dynamics of a given state from data, {\em i.e.} 
the time evolution of the variable of interest. Physical models usually describe this time devolution through an ordinary differential equation (ODE). One may distinguish two main families of data-driven approaches. A first category involves global parametric representations derived from physical principles  \cite{brunton_discovering_2016}. Polynomial representations are typical examples \cite{paduart_identification_2010}. The combination of such representations with sparse regression was recently shown to significantly open new research avenues. A second category of approach adopts a machine learning point of view and states the considered issue as a regression issue for a predefined time step $dt$, {\em i.e.} the regression of the state at time $t+dt$ given the state at time $t$. A variety of machine learning regression models have been investigated, among which neural networks and nearest-neighbor models (often referred to as analog forecasting models in geoscience) are the most popular ones \cite{zhao_analog_2014,lguensat_analog_2017}. Such approaches offer more modeling flexibility to optimize forecasting performance, at the expense however in general of a lack of interpretability of the learnt representation.

In this work, we investigate neural network (NN) representations for dynamical systems governed by some underlying but unknown ODEs. We aim to derive computationally-efficient and physically-sound representations. Our contribution is three-fold: i) we make explicit the interpretation of Runge-Kutta integration schemes as graphical models to introduce a residual NN architecture, ii) we introduce a NN architecture with bilinear layers to embed intrinsic non-linearities depicted by the dynamical systems, iii) we demonstrate the relevance of the proposed NN architecture with respect to state-of-the-art models both for model identification and forecasting for different classic systems, namely Lorenz-63 and Lorenz-96 dynamics \cite{lorenz_deterministic_1963}, which are representative of ocean-atmosphere dynamics, and Oregenator system  \cite{field_oscillating_1989}, which relates to oscillatory chemical dynamics. 

This paper is organized as follows. Section \ref{sec:Bi-RNN}
describes the proposed NN-based architecture for dynamical systems. Section \ref{s:exp} presents numerical experiments. We further discuss our contributions in Section \ref{s:disc}.

\section{Neural Net Architectures for Dynamical systems}
\label{sec:Bi-RNN}

We present in this section the proposed NN architectures to represent and forecast a dynamical system governed by an unknown ODE. We first point out the graphical representation of Runge-Kutta methods as residual neural nets. Based on this graphical representation, we introduce the proposed bilinear NN. 
We then discuss training issues and applications to forecasting and reconstruction problems.

\subsection{Runge-Kutta methods as residual neural nets}

Let us consider a dynamical system, whose time-varying state $X$ is governed by an ordinary differential equation (ODE):
\begin{equation}
\frac{dX_t}{dt}= M\left ( X_t, \theta \right )
\label{eq:sys dyn}
\end{equation}
where $\theta$ is some parameters. The fourth-order Runge-Kutta integration scheme is among the most classical ones for simulation state dynamics from a given initial condition $X(t_0)$. It relies on the following sequential update for a predefined integration time step $dt$:
\begin{equation}
\label{eq:rk_4}
X_{t_0+(n+1) dt} = X_{t_0+n \cdot dt} + \displaystyle \sum_{i=1}^4 \alpha_i k_i 
\end{equation}
$\{k_i\}$ are defined as follows : 
$k_i = M \left ( X_{t_0+\beta_i k_{i-1} dt},\theta \right )$
with $k_0=0$, $\alpha_1=\alpha_4=1/6$, $\alpha_2=\alpha_3=2/6$, $\beta_1=\beta_4=1$ and $\beta_2=\beta_3=1/2$.

Runge-Kutta integration scheme (\ref{eq:rk_4}) may be restated using a graphical model as illustrated in the bottom panel of Fig.\ref{fig:bi-NN-SL}. Assuming we know operator $F$, the fourth-order runge-Kutta scheme can be regarded as a recurrent network with a four-layer residual net \cite{lecun_deep_2015}, each layer sharing the same operator $F$. In this architecture, coefficients $\{\alpha_i\}_i$ refers to the relative weights given to the ouputs of the four repeated blocks $F$. The same holds for coefficient $\beta_i$ which refer to the weight given to the output from block $i-1$ when added to input $X_t$ when feeded to block $i$.

Based on this representation of numerical integration (\ref{eq:rk_4}) as a residual net, we may state the identification of dynamical operator $M$ in (\ref{eq:sys dyn}) as the learning of the parameters of block $F$ for a recurrent residual NN stated as in Fig.~\ref{fig:bi-NN-SL}. The other parameters, namely coefficients $\{\alpha_i\}_i$ and $\{\beta_i\}_i$, 
may be set to the values used in the fourth-order Runge-Kutta scheme or learnt from data. Overall, the key aspect of the consider residual recurrent net is the architecture and parameterization chosen for the shared block $F$. We may also stress that the fourth-order architecture sketched in Fig.\ref{fig:bi-NN-SL} may be extended to any lower- or higher-order scheme. As a special case, the explicit Euler scheme leads to a one-block architecture. 

Such NN representation of numerical schemes have been investigated in previous works \cite{anastassi_constructing_2014,tsitouras_neural_2002} to estimate coefficients $\{\alpha_i\}_i$ and $\{\beta_i\}_i$ for a given ODE. The objective is here to learn all the parameters of the NN representation of the unkown ODE governing observed time series.

\subsection{Proposed bilinear neural net architecture}

Neural net architectures classically exploit convolutional, fully-connected and non-linear activation layer \cite{lecun_deep_2015}. Following this classic framework, operator $F$ may be approximated as a combination of such elementary layers. It may be noted that dynamical systems, as illustrated for instance by Lorenz dynamics (\ref{eq:lorenz-63}) and (\ref{eq:lorenz-96}), involve non-linearities, which might not be well-approximated by the combination of a linear transform of the inputs and of a non-linear activation layer. Especially physical dynamical systems often involve bilinear non-linearities, which express some multiplicative interaction between two physical variables \cite{brunton_discovering_2016,mangan_model_2017}. Among classic physical models, one may cite for instance advection-diffusion dynamics or shallow water equations. Polynomial decompositions then appear as natural representation of dynamical systems for instance for model reduction issues \cite{paduart_identification_2010}.

\begin{figure}
\centering
\fbox{\includegraphics[width=5cm]{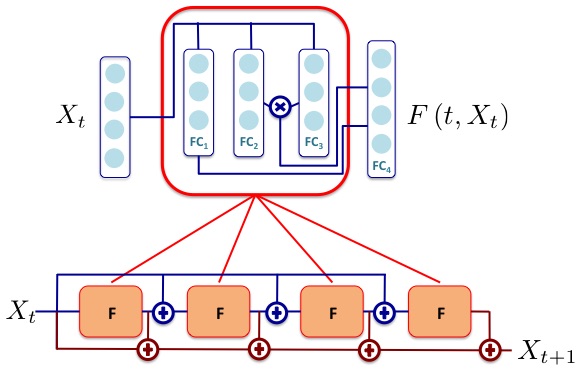}}
\caption{{{\bf   \em Proposed bilinear residual architecture for the representation of a dynamical system represented by (\ref{eq:sys dyn}).} We illustrate an architecture associated with a 4$^{th}$-order Runge-Kutta-like numerical integration for an elementary time-step $dt=1$. It involves a four-layer residual neural net with an elementary network $F$ repeated four times. The output of this elementary network involves a fully-connected layer $FC_4$ whose inputs are the concatenation of the output of the fully-connected layer $FC_1$ and the element-wise product between the outputs of fully-connected layers $FC_2$ and $FC_3$.} 
} 
\label{fig:bi-NN-SL}
\end{figure}

These considerations motivate the introduction of a bilinear neural net architecture. As illustrated in Fig.\ref{fig:bi-NN-SL}, we can combine fully-connected layers and an element-wise product operator to embed a second-order polynomial representation for operator $F$ in the proposed architecture. High-order polynomial representation might be embedded similarly. In Fig.\ref{fig:bi-NN-SL}, we illustrate an architecture where operator $F$ can be represented as the linear combination of three linear terms (i.e., linear combination of the input variables) and three bilinear terms (i.e., products between two linear combination of the input variables). In this architecture, the parameterization of the architecture initially relies on the definition of the number of linear and non-linear terms, which relate to the number of hidden nodes in the fully-connected layers, respectively $FC_1$ and $FC_{2,3}$. The calibration of the proposed architecture then comes to learning the weights of the different fully-connected layers. It may be noted that bilinear NN architectures have also been proposed in other context  \cite{lin_bilinear_2015,park_complex-bilinear_2002}.


\subsection{Training issues}


Given the proposed architecture and a selected parameterization, {\em i.e.} the number of nodes of the fully-connected layers $FC_{1,2,3}$ 
the number of $F$ blocks, the learning of the model aims primarily to learn the weights of the fully-connected layers associated with block $F$. As stated previously, coefficients 
$\{\alpha_i\}_i$ and $\{\beta_i\}_i$ from (\ref{eq:rk_4}) may be set {\em a priori} or learned from the data. Given a dataset $\{X_{t_n},X_{t_n+dt}\}_n$, corresponding to state time  series for a given time resolution $dt$, the loss function used for training is the root mean square error of the forecasting at one time step $dt$. Given the relationship between the number of elementary blocks $F$ in the considered architecture and the order of the underlying integration scheme, one may consider an incremental strategy, where we initially consider a one-block 
architecture, {\em i.e.} an explicit Euler integration scheme prior to increasing the number of $F$-blocks for a higher-order numerical scheme.

Regarding initialization aspects, the weights of the fully-connected layers $FC_{1,2,3,4}$ 
are set randomly and coefficients $\{\alpha_i\}_i$ and $\{\beta_i\}_i$ are set to those of the associated Runge-Kutta scheme. 
We use Keras framework with Tensorflow backend to implement the proposed architecture. During the learning step, we impose a hard constraint that the different $F$-blocks share the same parameters after each training epoch. 

\subsection{Application to forecasting and latent dynamics identification}

In this study, we first consider forecasting of the evolution of state $X$ from a given initial condition $X_{t_0}$.
For a trained NN architecture, two strategies may be considered: i) the use of the trained architecture as a recurrent neural net architecture to forecast a time series for a number of predefined time steps $dt$, ii) the plug-an-play use of the trained operator $F$ in a classic ordinary differential equation solver. It may be noted that, for a trained operator $F$, the fourth-order architecture sketched in Fig.~\ref{fig:bi-NN-SL} is numerically equivalent to a fourth-order Runge-Kutta solver. 

We also explore the potential of the proposed NN representation for the identification of latent dynamics. More specifically, we assume that we are provided with a time series of states $\{Y_{t+kdt}\}_k$, which is driven by the dynamics of a  latent lower-dimensional state $X_t$ according to a linear mapping, {\em i.e.} $Y_{t+kdt}=HX_{t+kdt}$ with $H$ a linear mapping from the low-dimensional space to the hig-dimensional one. To address the identification of the dynamics of latent state $X_t$ from time series $\{Y_{t+kdt}\}_k$, we consider the proposed NN representation in which $FC_1$ layer serves as a mapping layer from the high-dimensional space to the low-dimensional one and $FC_4$ layer becomes mapping block from the low-dimensional block to the high-dimensional one.

\section{Numerical experiments}
\label{s:exp}

This section presents the numerical experiments we perform to demonstrate the relevance of the proposed bilinear NN architecture. We introduce the considered case-studies and our experimental setup, and report results including the benchmarking with respect to state-of-the-art schemes.

\subsection{Considered case-studies and experimental setup}
We consider three reference dynamical systems to perform a qualitative and quantitative evaluation of the proposed architecture: namely, Lorenz-63, Oregonator and Lorenz-96 dynamics. For all models, we generate time series exemplars from the numerical integration of the ODE which governs each system. We use the Shampine and Gordon solver \cite{ashino_behind_2000}.

The Lorenz-63 system is a 3-dimensional system governed by the following ODE:
\begin{equation}
\label{eq:lorenz-63}
\left \{\begin{array}{ccl}
\frac{dX_{t,1}}{dt} &=&\sigma \left (X_{t,2}-X_{t,2} \right ) \\
\frac{dX_{t,2}}{dt}&=&\rho X_{t,1}-X_{t,2}-X_{t,1}X_{t,3} \\
\frac{dX_{t,3}}{dt} &=&X_{t,1}X_{t,2}-\beta X_{t,3}
\end{array}\right.
\end{equation}
Under parameterization $\sigma =10$, $\rho=28$ and  $\beta=8/3$, Lorenz-63 system involves chaotic dynamics with two attractors. The integration time step $dt$ is set to 0.01.   

Oregonator system is also a 3-dimensional system. This stiff dynamical system is governed by: 
\begin{equation}
\left \{
\begin{array}{ccl}
\frac{dX_{t,1}}{dt}&=&\alpha\left(X_{t,2}+X_{t,1}(1-\beta X_{t,1}-X_{t,2})\right) \\
\frac{dX_{t,2}}{dt}&=&\frac{1}{\alpha}\left(X_{t,3}-(1+X_{t,1})X_{t,2} \right ) \\
\frac{dX_{t,3}}{dt}&=&\sigma\left(X_{t,1}-X_{t,3}\right)\\
\end{array}\right.
\end{equation}
Here, we consider $\alpha=77.27$, $\beta=8.375.10^{-6}$ and $\sigma =0.161$. The integration time step $h$ is set to 0.1.

Lorenz-96 system is a 40-dimensional system. It involves propagation-like dynamics governed by:
\begin{equation}
\label{eq:lorenz-96}
\frac{dX_{t,i}}{dt}=(X_{t,i+1}-X_{t,i-2})X_{t,i-1}+A
\end{equation}
with periodic boundary conditions ({\em i.e.} $X_{t,-1}=X_{t,40}$ and $X_{t,41}=X_{t,1}$). Time step $h$ is set to 0.05 and $A=9$.

For each system, we generate a time series of 50000 time steps to create our training dataset and a time series of 1000 time steps for the test dataset. For a given data-driven representation, we evaluate the forecasting performance as the Root Mean Square error (RMSE) for an integration time step of $h$, 4$h$ and 8$h$, where $h$ is the integration time step of the simulated time series. The RMSE is averaged over all the initial conditions taken from the test time series. For benchmarking purposes, we compare the proposed bilinear residual NN representation to the following data-driven representation:
\begin{itemize}
\item a sparse regression model \cite{brunton_discovering_2016} referred to as SR. It combines an augmented bilinear state as regression variable and a sparsity-based regression; 
\item an analog forecasting operator \cite{lguensat_analog_2017} referred to as AF. It applies locally-linear operators estimated from nearest neighbors, retrieved according to a Gaussian kernel as in \cite{lguensat_analog_2017};
\end{itemize}
Several NN representations are evaluated:  
\begin{itemize}
\item the proposed bilinear residual architecture using a one-block version (Euler-like setting), referred to as Bi-NN(1), and a four-block version (Runge-Kutta-like setting) with share layers, referred to as  Bi-res-NN-SL(4). We use  3-dimensional (resp. 40-dimensional) fully-connected layers for the linear and bilinear layers $FC_{1,2,3}$ for Lorenz-63 and Oregonator systems (resp. Lorenz-96 system).
\item a neural network architecture similar to the above four-block one but replacing the proposed bilinear block by a classic MLP. From cross-validation experiments, we consider a MLP with 5 hidden layers (resp. 11 hidden layers) and 6 nodes in each layer (resp. 80 nods in each layer) for both Lorenz-63 and Oregonator models (resp. Lorenz-96 model). This architecture is referred to as MLP-SL(4);
\item a MLP architecture trained to predict directly state at time $t+h$ from the state at time $t$. From cross-validation experiments, we consider a MLP with 5 hidden layers (resp. 10 hidden layers) and 6 nodes in each layer (resp. 80 nodes) for the Lorenz-63 and the Oregonator models (resp. the Lorenz-96 model). This architecture is referred to as MLP.
\end{itemize}



\subsection{Results}

~\\{\bf Learning from noise-free training data:} in this experiment, we compare the quality of the forecasted state trajectories generated using the models described above. The learning of the data-driven models is carried using noise-free time series computed using the analytical dynamical models.

\begin{table}[hbpt]
\label{tab:forec1}
\caption {{\bf  \em Forecasting performance of data-driven models for Lorenz-63, Oregonator and Lorenz-96 dynamics}: mean RMSE for different forecasting time steps for the floowing models,  AF (A), SR (B), MLP (C), MLP-SL(4) (D), Bi-NN(1) (E), Bi-NN-SL(4) (F). See the main text for details.}

\centering
\begin{adjustbox}{max width=0.455\textwidth}
\begin{tabular}{l*{6}c}
\toprule
 & A & B & C & D & E & F\\
\midrule \midrule 
Lorenz-63\\
$t_0+h$  & $0.001$ & $0.002$ & $0.114$ & $0.009$ & $0.002$ & \bf 1.37E-5\\
$t_0+4h$ & $0.004$ & $0.008$ & $0.172$ & $0.035$ & $0.006$ & \bf 4.79E-5\\
$t_0+8h$ & $0.007$ & $0.014$ & $0.197$ & $0.071$ & $0.013$ & \bf 8.17E-5\\                        
\midrule 
Oregonator\\
$t_0+h$  & $>10^2$ & $6.921$ & $3.159$ & $4.503$ & \bf 0.035 & 0.038 \\
$t_0+4h$ & $>10^2$ & $7.558$ & $4.660$ & $4.961$ & $4.458$ & \bf 4.296 \\
$t_0+8h$ & $>10^2$ & $8.275$ & $4.268$ & $4.249$ & $3.524$ & \bf 3.448 \\
\midrule 
Lorenz-96\\
$t_0+h$  & $0.242$ & $0.031$ & $0.827$ & $0.731$ & $0.049$ & \bf 0.012\\
$t_0+4h$ & $0.580$ & $0.086$ & $1.623$ & $1.870$ & $0.140$ & \bf 0.035\\
$t_0+8h$ & $0.988$ & $0.147$ & $2.215$ & $2.752$ & $0.246$ & \bf 0.064\\
\bottomrule
\end{tabular}
\end{adjustbox}
\end{table}

~\\{\bf Model identification:} We  investigate model identification performance  for Lorenz-63 dynamics in Tab.\ref{tab: res1}. We report  the performance in terms of model parameter estimation for the three data-driven schemes whose parameterization explicitly relates to the true physical equations, namely SR, Ni-NN(1) and Bi-NN(4)-SL. Bi-NN(4)-SL leads to a better estimation of model parameters, which explains the better forecasting performance.



\begin{table}[H]
\caption {{\bf \em MSE in the estimation of Lorenz-63 parameters
for SR, Bi-NN(1) and Bi-NN(4)-SL models.} See the main manuscript for details.}
\label{tab: res1}
\centering
\begin{tabular}{l c c c}
\toprule
&SR          & Bi-NN(1)    &Bi-NN(4)-SL\\
\midrule \midrule
MSE&0.0387 & 0.2570 & \bf 0.0239\\
\bottomrule
\end{tabular}
\end{table}

\begin{figure}
\centering
\fbox{\includegraphics[width=7cm]{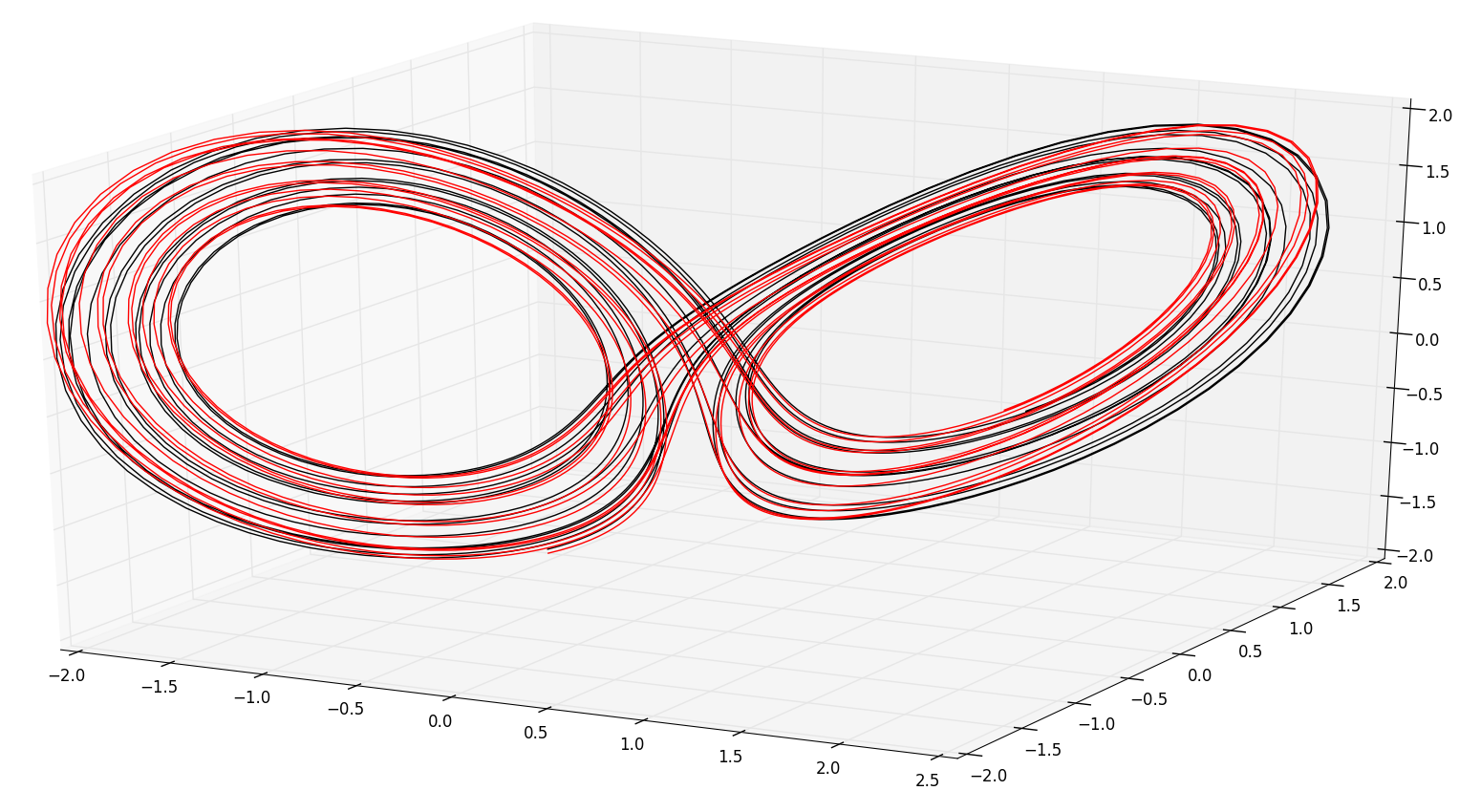}}
\caption{{\bf \em Identification of the latent Lorenz-63 dynamics of an observed 5-dimensional dynamical system:} we illustrate the reconstructed 3-dimensional latent dynamics (black,-) w.r.t. the true one (red,-) using the proposed Bi-NN(1) model. See the main text for details.} 
\label{fig:projatt}
\end{figure}

~\\{\bf Identification of latent Lorenz-63 dynamics:}
We further illustrate the potential of the proposed bilinear NN representation for the identification of latent lower-dimensional dynamics for an observed dynamical system. In Fig.\ref{fig:projatt}, we consider a 5-dimensional system, driven by underlying Lorenz-63 
dynamics according to a linear mapping. Using solely a time series of observations of the 5-dimensional system with $dt=0.01$, we successfully identify the underlying low-dimensional chaotic behavior. It may be noted that the identification issue is achieved up to a 3$\times$3 rotation matrix.

\section{Conclusion}
\label{s:disc} 
 
In this work, we demonstrated the relevance of a residual bilinear neural net representation for the modeling and identification of dynamical systems. Our NN-based representation relies on the representation of classic numerical scheme as a multi-layer network.  This NN representation opens new research avenues for the exploitation of machine-learning-based and physically-sound strategies for the modeling, identification and reconstruction of dynamical systems.


\vfill\pagebreak

\bibliographystyle{IEEEbib}
\bibliography{Zotero}

\end{document}